\documentclass{article}

\usepackage[preprint]{neurips_2020}
\usepackage[utf8]{inputenc} 
\usepackage[T1]{fontenc}    
\usepackage{hyperref}       
\usepackage{url}            
\usepackage{booktabs}       
\usepackage{amsfonts}       
\usepackage{nicefrac}       
\usepackage{microtype}      
\usepackage{caption}
\usepackage{subcaption}
\usepackage{graphicx}
\usepackage{amsmath}
\usepackage{multirow}
\usepackage{todonotes}
\usepackage[ruled,vlined]{algorithm2e}

\title{Does Data Augmentation Benefit from \\ Split BatchNorms?}

\author{%
  Amil Merchant\thanks{Work done as a member of the Google AI Residency program \url{g.co/airesidency}} \\
  Google Research \\
  Mountain View, CA 94043 \\
  \texttt{amilmerchant@google.com} 
\And
Barret Zoph \\
  Google Brain \\
  Mountain View, CA 94043 \\
  \texttt{barretzoph@google.com}
\And
Ekin Dogus Cubuk \\ 
  Google Brain \\
  Mountain View, CA 94043 \\
  \texttt{cubuk@google.com}
}

\begin{document}

\maketitle

\begin{abstract}
    Data augmentation has emerged as a powerful technique for improving the performance of deep neural networks and led to state-of-the-art results in computer vision. However, state-of-the-art data augmentation strongly distorts training images, leading to a disparity between examples seen during training and inference. In this work, we explore a recently proposed training paradigm in order to correct for this disparity: using an auxiliary BatchNorm for the potentially out-of-distribution, strongly augmented images. Our experiments then focus on how to define the BatchNorm parameters that are used at evaluation. To eliminate the train-test disparity, we experiment with using the batch statistics defined by clean training images only, yet surprisingly find that this does not yield improvements in model performance. Instead, we investigate using BatchNorm parameters defined by \textit{weak} augmentations and find that this method significantly improves the performance of common image classification benchmarks such as CIFAR-10, CIFAR-100, and ImageNet. We then explore a fundamental trade-off between accuracy and robustness coming from using different BatchNorm parameters, providing greater insight into the benefits of data augmentation on model performance. 
\end{abstract}

\section{Introduction}

Data augmentation has become a common technique for improving the diversity of examples within datasets for machine learning without needing to explicitly label additional examples. The benefits of this strategy have been seen in a number of application domains, including image classification \cite{simard2003best, krizhevsky2012imagenet, Devries_Improved, Zhang_Mixup}, object detection \cite{girshick2018detectron}, and semantic segmentation \cite{fang2019instaboost}. With respect to images, common examples of data augmentations include MixUp \cite{Zhang_Mixup}, Cutout \cite{Devries_Improved}, CutMix \cite{Yun_CutMix}, and various forms of Gaussian noise \cite{lopes2019improving}. A more diverse set of transformations (such as those from the PIL library\footnote{https://pillow.readthedocs.io/en/5.1.x}) has been utilized by Ratner and Ehrenberg et al.~\cite{ratner2017learning}, AutoAugment~\cite{Cubuk_Autoaugment,zoph2019learning}, and RandAugment~\cite{Cubuk_Randaugment}. Such data augmentation strategies have been used to achieve state-of-the-art results in image classification~\cite{Cubuk_Autoaugment,Cubuk_Randaugment,Zhang_Adversarial,Xie_Adversarial}, object detection~\cite{zoph2019learning,cheng2020improving}, semi-supervised learning~\cite{xie2019unsupervised,xie2020self,zoph2020rethinking}, contrastive learning~\cite{khosla2020supervised}, and robustness to test-time image distortions~\cite{Yin_Fourier,Xie_Adversarial,xie2020self,hendrycks2019augmix}.     

\begin{figure}
    \begin{subfigure}{0.19\linewidth}
        \centering
        \includegraphics[width=\linewidth]{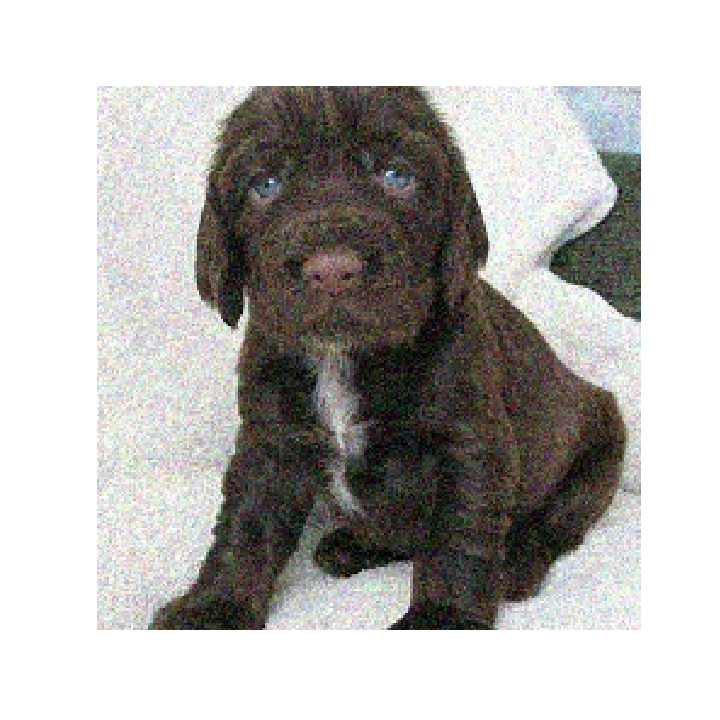}
        \caption{Clean Image}
    \end{subfigure} %
    \begin{subfigure}{0.19\linewidth}
        \centering
        \includegraphics[width=\linewidth]{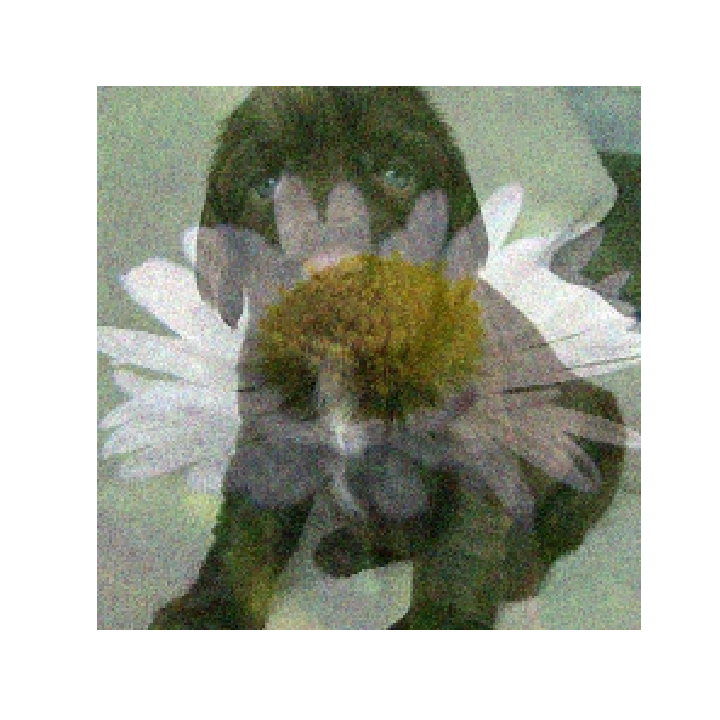}
        \caption{MixUp}
    \end{subfigure} %
    \begin{subfigure}{0.19\linewidth}
        \centering
        \includegraphics[width=\linewidth]{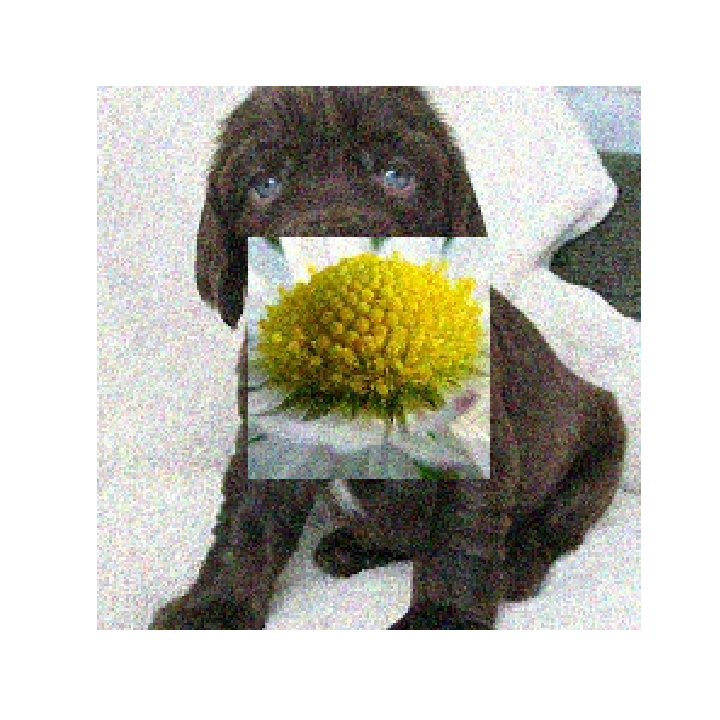}
        \caption{CutMix}
    \end{subfigure} %
    \begin{subfigure}{0.19\linewidth}
        \centering
        \includegraphics[width=\linewidth]{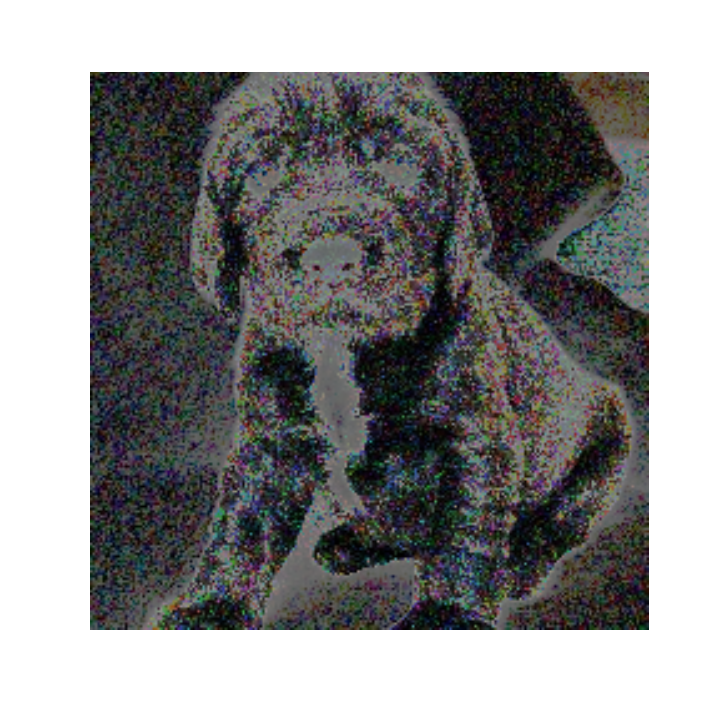}
        \caption{AutoAugment}
    \end{subfigure}
    \begin{subfigure}{0.19\linewidth}
        \centering
        \includegraphics[width=\linewidth]{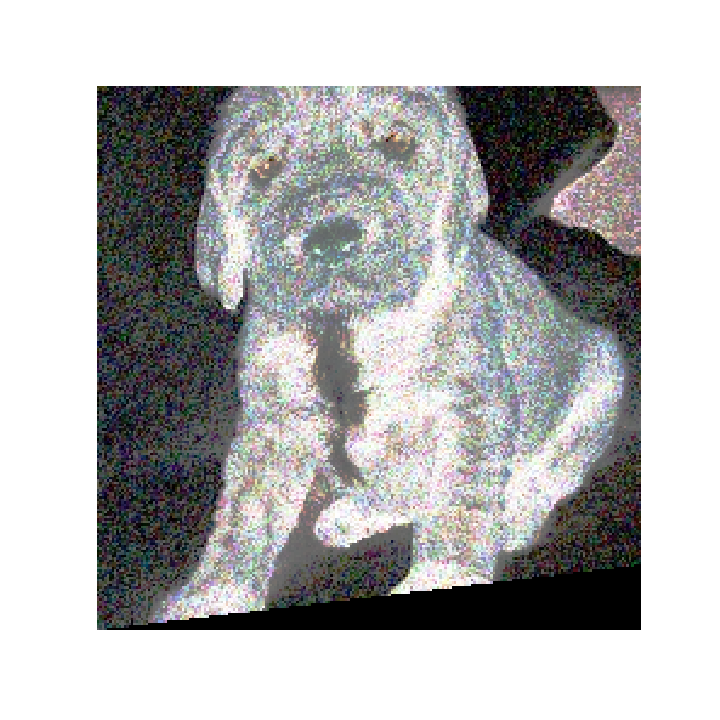}
        \caption{RandAugment}
    \end{subfigure}
    \centering
    \caption{Examples of images produced by a number of common data augmentation techniques. The produced images often appear unnatural, and the examples above highlight the disparity between examples seen during training and the clean (un-augmented) examples seen at test-time.}
    \label{fig:augmentations}
\end{figure}

Early work in data augmentation often assumed that beneficial techniques would produce images that would be close to the true data distribution \cite{bellegarda1992robust,simard2003best}. However, with many of the techniques above, it is clear that the resulting images are unnatural and likely to be out-of-distribution with respect to the test set (see Figure \ref{fig:augmentations} for examples). Images with augmentations are often blended together or modified so strongly that the image semantics are destroyed. 

Therefore, despite the improvements in model performance from these techniques, there still exists a train-test disparity in the resulting model. A number of recent works have attempted to adjust for this incongruity, suggesting that models can be improved by the early stopping of various data augmentation strategies \cite{he2019data, gontijo2020affinity} or by density matching between the clean (un-augmented) and augmented datasets \cite{lim2019fast, hataya2019faster}.
However, despite the corrections, these methods have not outperformed augmentation strategies that heavily distort the training images \cite{Zhang_Mixup, Yun_CutMix, Cubuk_Randaugment}.

Instead, this train-test disparity can also be addressed by correcting the parameters in BatchNorm layers, which are commonly thought to capture distributional information about input images. To do so, we adopt a recently proposed training paradigm where an auxiliary BatchNorm is used for the potentially out-of-distribution (OOD) augmented images \cite{Wang_Transferable, zajkac2019split, Xie_Adversarial}. This technique re-aligns the BatchNorm parameters to match the distribution of representations seen at inference. For example, Transfer Normalization achieved state-of-the-art results on a number of domain transfer tasks where there is an explicit train-test gap \cite{Wang_Transferable}.

Applying this technique to data augmentation however raises a number of questions. For example, how should the BatchNorm parameters used at evaluation be defined? Do the statistics defined by clean images (compared to strongly augmented images) yield improved performance? What happens to the robustness properties of the network from using this separated BatchNorm setup? Our findings from these questions can be summarized as follows:

\begin{itemize}
    \itemsep0em
    \item Naively separating all clean (un-augmented) and augmented images into two separate BatchNorms during training is often harmful for model performance. We hypothesize that this lack of improvement occurs as having a portion of the training images being un-augmented leads to overfitting (Section \ref{sec:clean}).
    \item Using \textbf{weak augmentations} significantly improves the performance of the separated BatchNorm setup on multiple image classification benchmarks. We then explore how this strategy relates to proposed measures for in-distribution data augmentations (Section \ref{sec:weak}). \cite{gontijo2020affinity}. Section \ref{sec:ablation} provides ablation studies to locate the improvements from weak augmentations.
    \item Analyzing the robustness properties of using a separated BatchNorm presents a fundamental trade-off between improvements in accuracy and corruption error, providing a clearer picture on the effects of data augmentation (Section \ref{sec:robustness}).
\end{itemize}

\section{Related Work}
\paragraph{Data Augmentation}
Early examples of data augmentations often focused on creating realistic but `different` training examples \cite{bellegarda1992robust} and often included horizontal flips, crops, and minor color distortions to MNIST and CIFAR-10 images \cite{ciregan2012multi, sato2015apac, simard2003best}. However, this is clearly no longer the case in modern models as data augmentation techniques that produce out-of-distribution and heavily modified images have been shown to significantly improve performance \cite{bengio_deep, lopes2019improving, zhong2017random, Zhang_Mixup}.

Based on similar observations, a number of recent studies have shown interest in understanding the effects of data augmentation and the relation to the true underlying data distribution. For example, Gontijo-Lopes et al. \cite{gontijo2020affinity} proposed the Affinity metric to measure how in-distribution a given data augmentation is. The method is based on the evaluation performance of a clean model tested on augmented data. Their work found that beneficial augmentations can often be out-of-distribution. On the other hand, techniques such as AugMix have also shown success correcting for the distributional shift in data augmentation by producing more natural-looking outputs. Similarly, Fast AutoAugment and Faster AutoAugment \cite{lim2019fast, hataya2019faster} are both AutoML-based techniques and use core ideas from density matching to minimize the distance between the augmented and clean data. Taken together, these studies suggest that data augmentation faces two competing effects: improvements coming from diverse examples and producing outputs that match the data distribution.

\paragraph{Importance of BatchNorm Statistics}
In this work, we choose to address the train-test disparity coming from augmentation by correcting the BatchNorm layers, which are often thought to capture domain specific effects during training \cite{Ioffe_Batch}. A number of recent works have highlighted the importance of aligning these parameters to the test domain. For example, in the field of domain transfer, AdaBN \cite{li2016revisting} and AutoDIAL \cite{cariucci2017autodial} have shown that recomputing BatchNorm parameters based on the test domain can improve model performance. Transfer Normalization \cite{Wang_Transferable} introduced the idea of separated batch statistics for different domains and designed an end-to-end trainable layer that achieved state-of-the-art performance on a number of domain transfer benchmarks.

\paragraph{Seperated BatchNorm Layers} This idea of using separated BatchNorms for potentially out-of-distribution data has been adopted by a number of application areas with promising performance. In the field of semi-supervised learning, separated BatchNorm parameters allowed models to better incorporate unlabeled images that did not correspond to any of the labeled classes \cite{zajkac2019split}. Most closely related to our work is AdvProp \cite{Xie_Adversarial}, which found that strong adversarial noise could be incorporated into image classification models through the use of an auxiliary BatchNorm. EfficientNet \cite{tan2019efficientnet} models trained with this approach significantly improved results on the ImageNet \cite{krizhevsky2012imagenet} and ImageNet-C benchmarks \cite{Hendrycks_Benchmarking}. While their work showed impressive performance gains, there still remain a number of open questions regarding the use of separated BatchNorms in the context of data augmentation. For example, are the batch statistics defined by clean images optimal? How does the use of separated BatchNorms impact model robustness?

\section{Methods}
\begin{figure}
\centering
\begin{minipage}{0.35\textwidth}
    \includegraphics[scale=0.4]{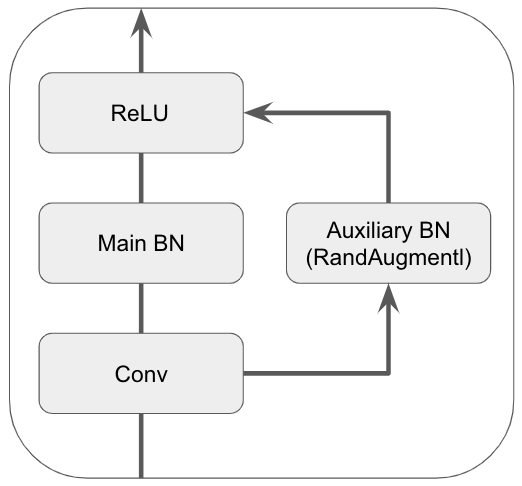}
\end{minipage}%
\begin{minipage}{0.05\textwidth}
\end{minipage}%
\begin{minipage}{0.60\textwidth}
\begin{algorithm}[H]
\label{alg:advprop}
\KwData{mini-batch of images $x^c$}
Produce augmented batch $x_m$ for main branch\;
Produce augmented batch $x_a$ for augmented branch\;
Compute loss $L^m(x^m)$ using main BN parameters\;
Compute loss $L^a(x^a)$ using auxiliary BN parameters\;
Compute  total loss as $L_{total} = (L^m + L^a) / 2$\;
Update the parameters with respect to $L_{total}$\;
\caption{Single Step with Separated BatchNorms}
\end{algorithm}
\end{minipage}
    \caption{A schematic diagram (left) and explanation of a single update step (right) for a model with separated BatchNorms. Note how in our setup, the strong data augmentation techniques (e.g. RandAugment, Gaussian Noise) are placed onto the auxiliary BN parameters. In this paper, we vary the augmentations for the main branch, ranging from no augmentation to RandAugment.}
    \label{fig:advprop}
\end{figure}

\paragraph{Separate BatchNorms} 
In order to train models with the separated BatchNorm setup, this paper follows prior work \cite{Wang_Transferable}; in particular, the setup can be best thought of as a simplified version of fine-grained AdvProp\footnote{Note, AdvProps studies a specific case of the generic setup of separated BatchNorms, where AutoAugment is applied to all images and adversarial noise is limited to the auxiliary BatchNorm.} \cite{Xie_Adversarial}. The associated training strategy first copies the input mini-batch and separately applies noise or augmentation techniques to each. One set is given to the main branch (with a specific set of BatchNorm parameters), whereas the potentially out-of-distribution images utilize an auxiliary BatchNorm. During training, the losses from these two branches are averaged, whereas evaluation is only performed using the batch statistics from the main branch (as in Figure \ref{fig:advprop}).

\paragraph{Augmentation on the Auxiliary BatchNorm} As a popular and effective benchmark for data augmentation, RandAugment serves as a reasonable starting point for analyzing the separate BatchNorm setup\footnote{RandAugment assume the default augmentations of flips, crops, and Cutout are included, as per Cubuk et al.\cite{Cubuk_Randaugment}} \cite{Cubuk_Randaugment}. This augmentation strategy combines over 15 individual augmentation types, including shears, rotations, and color equalization. For each image, two or three of these augmentations are applied at a user-defined augmentation strength. This method provides a particularly promising starting point as RandAugment has led to state-of-the-art performance on a number of image classification tasks, yet little to no effort has been put in to ensure that the outputs are \textit{natural} and match the test distribution. The analysis for other common data augmentation techniques is provided in Section \ref{sec:extensions}.

\paragraph{Datasets and Models} 
Experiments are provided for three common image classification benchmarks. Two datasets, CIFAR-10 and CIFAR-100 \cite{krizhevsky2009learning}, consist of tiny-images containing 50K training and 10K test images apiece. Our analysis uses WideResNet-28-2 and WideResNet-28-10 \cite{zagoruyko2016wide} models that are trained for 200 epochs with a learning rate of 0.1, batch size of 128, weight decay of 5e-4, and cosine learning rate decay. To ensure that the observed trends extend to larger datasets, experiments are also conducted on ImageNet \cite{deng2009imagenet}, which contains approximately 1.2 million colored images. The corresponding ResNet50 models follow default hyper-parameters \cite{Cubuk_Randaugment} and are trained for 180 epochs using an image size of 224x224. The models use a weight decay of 1e-4, a momentum optimizer with parameter value of 0.9, a batch size of 4096. and a learning rate of 0.1 (scaled by the batch size divided by 256).

\paragraph{Metrics} In general, top-1 accuracy is reported for the clean test set, but for ImageNet top-5 numbers are also included. All metrics for CIFAR-10 and CIFAR-100 are reported from an average of 10 training runs. Robustness results are provided utilizing the Common Corruptions benchmark \cite{Hendrycks_Benchmarking} of CIFAR-10-C and ImageNet-C, which provides test data with 15 different types of inputted noise at 5 different intensities each. The error rate for a corruption $c$ and severity $s$ is given by $E_{c,s}$. For CIFAR-10-C, the associated metric is the un-normalized corruption error ($uCE_c$), whereas for ImageNet-C, the robustness metrics are normalized by the corruption error of AlexNet \cite{krizhevsky2012imagenet} ($CE_c$). Averages over all corruptions are reported. Note that in all cases, lower corruption error is better. 
$$uCE_{c} = \frac{1}{5} \sum_{s=1}^5 E_{c,s} \qquad CE_c = \frac{\sum_{s=1}^5 E_{c,s}}{\sum_{s=1}^5 E_{c,s}^{\text{AlexNet}}}$$

\section{Naively Using Separated BatchNorms Yields No Improvement}
\label{sec:clean}
If the strongly augmented images use the auxiliary BatchNorm during training, a fundamental question becomes how to define the batch statistics that are used for evaluation (those on the main branch). A naive strategy would be to use clean images only, with the expectation that the clean images would best match the ``test data''. However, the resulting CIFAR-10 model trained on Wide-ResNet-28-2 displayed significant degradation in performance, with a decrease of 1.6\% in accuracy compared to a model trained on RandAugment only (as shown in top row of Table \ref{tab:affinity}). This diminished performance likely arises from the lack of diversity when defining the batch statistics for evaluation and possible over-fitting to the training set.

To counter this effect, we devise a simple change to the separated BatchNorm setup. Instead of using clean images directly, incorporating simple \textit{weak augmentations} yields significant gains in performance. This strategy is similar to the recently proposed methods in Semi-Supervised Learning such as FixMatch \cite{sohn2020fixmatch} yet novel in it's application to multiple BatchNorms. We refer to this method in the text as Weak Augment.

\begin{table}[ht!]
    \centering 
    \begin{tabular}{l|r|r}
        \multicolumn{3}{c}{\textbf{Performance from using Weak Augmentations}} \\
        \hline 
        \textbf{CIFAR-10} & Affinity & $\Delta$ Clean Test Accuracy \\
        None & 0.0 &  -1.6 \\
        Flip & -0.8 & 0.1 \\
        Flip + Crop & -2.7 & 0.1 \\
        \textbf{Cutout} & \textbf{-16.1} & \textbf{0.5}\\
        AugMix & -12.7 & 0.3 \\
        Gaussian ($\sigma = 0.2$) & -25.7 & -0.3 \\
        RandAugment & -25.0 & 0.0 \\
        \multicolumn{2}{c}{} \\ 
    \end{tabular}
    \caption{Delta ($\Delta$) clean test accuracy for a number of augmentation strategies on the main BatchNorm, where RandAugment is applied on the auxiliary BatchNorm. The baseline is a model trained only with RandAugment without separate BatchNorms (accuracy of 95.8\%). Our found optimal weak augmentation (ex: Cutout) improves accuracy significantly on these CIFAR-10 WideResNet-28-2 models. Interestingly, we note that this is not the most in-distribution augmentation, as measured by Affinity \cite{gontijo2020affinity}.}
    \label{tab:affinity}
\end{table}

\section{Weak Augmentations Allow Separated BatchNorms to Be Effective}
\label{sec:weak}
To test for this hypothesized effect, a variety of weak augmentations were applied in the separated BatchNorm setup for WideResNet-28-2 models on CIFAR-10. The results presented in Table \ref{tab:affinity} show that incorporating the standard flip or crop augmentations into the main BatchNorm can recover the performance of a model trained only with RandAugment that is not using Separated BatchNorms. Applying slightly stronger weak augmentation in the form of Cutout \footnote{Cutout is performed in addition to flip and crop augmentations and is performed with a pad-size of 16/90 for CIFAR-10/ImageNet models respectively.} leads to significant gains in model performance of +0.5 on the CIFAR-10 dataset. Interestingly, note that this finding does not necessarily coincide with the current understanding of in-distribution augmentations and BatchNorms. The provided Affinity metric \cite{lopes2019improving} quantifies the distribution shift arising from a given augmentation by testing a model trained only on clean images on augmented images. Cutout is clearly not an exact-match for the true data distribution, yet empirically performs best in the experimental trials.

Nevertheless, this weak augmentation strategy devised on CIFAR-10 can be extended to a variety of benchmark tasks, with results provided in Table \ref{tab:clean}. Weak augmentations show significant gains in performance across the board for the CIFAR-10, CIFAR-100, and ImageNet benchmarks, over the already strong baselines set by RandAugment.

\begin{table}
    \centering
    \begin{tabular}{l|rr|rr|r}
        \multicolumn{6}{c}{\textbf{Clean Test Accuracy From Using Weak Augmentations}} \\
        \hline
        & Standard & Cutout & AA & RA & Weak Augment \\
        \hline
        \textbf{CIFAR-10} & & & & & \\ 
        Wide-ResNet-28-2 & 93.8 & 94.9 & 95.9 & 95.8 & \textbf{96.3} $\pm$ \textbf{0.1} \\
        Wide-ResNet-28-10 & 95.5 & 96.6 & 97.4 & 97.3 & \textbf{97.6} $\pm$ \textbf{0.1} \\
        \hline 
        \textbf{CIFAR-100} & & & & & \\
        Wide-ResNet-28-2 & 70.9 & 75.4 & 78.5 & 78.3 & \textbf{79.2} $\pm$ \textbf{0.2} \\
        Wide-ResNet-28-10 & 78.8 & 81.2 & 82.9 & 83.3 & \textbf{83.8} $\pm$ \textbf{0.3} \\
        \hline
        \textbf{ImageNet (Top-1/Top-5)} & & & & & \\
        ResNet50 & 76.3/93.1 & -- & 77.6/93.8 & 77.6/93.8 & \textbf{77.9/93.9} \\
        \multicolumn{6}{c}{} \\ 
    \end{tabular}
    \caption{Clean accuracy for the CIFAR-10 and CIFAR-100, including Standard (horizontal flips and random crops), Cutout \cite{Devries_Improved}, AutoAugment (AA) \cite{Cubuk_Autoaugment}, RandAugment (RA) \cite{Cubuk_Randaugment}, and the newly proposed weak augmentation setup. Weak Augment is defined by RandAugment on the auxiliary BatchNorm and Cutout on the main BatchNorm. All metrics are provided as the average over 10 runs.}
    \label{tab:clean}
\end{table}

\subsection{Ablations Locate Improvements from Weak Augmentions}
\label{sec:ablation}
While the results from \textit{weak augmentations} are impressive, the results on the benchmark tasks do not fully elucidate where the benefits of the training strategy are coming from. Table \ref{tab:ablations} compares model performance when procedurally adding the components of the separated BatchNorm setup. The use of two stochastic applications of RandAugment on each training batch does not change test performance (Table \ref{tab:ablations}, row 3), suggesting that the simple addition of data during training is not impacting model performance (contrasting from prior work \cite{hoffer2019augment}). Similarly, the use of a separated BatchNorm when the augmentations applied to the two branches do not differ does not yield any improvements (Table \ref{tab:ablations}, row 4).

Instead, the gains in model performance arise partially from the inclusion of \textit{weak} augmentations even without utilizing separate BatchNorms, improving by $0.3\%$ (Table \ref{tab:ablations}, row 5). Note, using separated BatchNorm layers with shared $\beta$, $\gamma$ parameters (but differing moving means and variances) shows no additional improvement in performance, but two fully independent BatchNorms increases accuracy by $0.2\%$ (Table \ref{tab:ablations}, row 7). These findings suggest novel insights into the effectiveness of separated BatchNorms. First, part of the benefit comes from the additional diversity of examples. Beyond that, additional improvements do not arise from having separate mean and variance moving averages.
This result is interesting as these moving averages are often thought to capture domain effects \cite{li2016revisiting}, yet in our experiments correcting these mean and variance does not yield improvements. Instead, two fully independent BatchNorms with separate $\beta$ and $\gamma$ are required.

\begin{table}
    \centering 
    \begin{tabular}{l|r}
        \multicolumn{2}{c}{\textbf{Ablation Study for Clean Test Accuracy}} \\ 
        \hline
        \textbf{CIFAR-10} & Accuracy \\ 
        (1) Baseline (Flips and Crops) & 94.9 \\
        (2) RandAugment & 95.8 \\
        (3) Two RandAugment Batches & 95.8 \\ 
        (4) Two RandAugment Batches with separated BatchNorm & 95.8 \\
        (5) Weak Augment without separated BatchNorms & 96.1 \\
        (6) Weak Augment with shared $\beta$, $\gamma$ parameters & 96.1 \\
        (7) Weak Augment & 96.3 \\
        \multicolumn{2}{c}{} \\ 
    \end{tabular}
    \caption{Ablation study for the improvements coming from weak augmentations, evaluated using a WideResNet-28-2 model on CIFAR-10. Relative to a RandAugment model, the addition of an extra batch of strongly augmented data does not improve model performance (3). The direct application of an auxiliary BatchNorm without having differing augmentations also yields constant performance (4). The benefits from Weak Augment come first from the inclusion of weakly augmented images (5) and then again from using two separate BatchNorms (7).}
    \label{tab:ablations}
\end{table}

\subsection{Experimenting with Other Auxiliary BatchNorm Data Augmentation Types}
\label{sec:extensions}
The Weak Augment strategy is generic and extends to a variety of other augmentation setups. Specifically, the strong augmentation provided on the auxiliary BatchNorm does not need to be RandAugment. In Table \ref{tab:extensions}, the results show that \textit{weak augmentations} are effective at improving the performance for a variety of data augmentation methods on the auxiliary BatchNorm. In Table \ref{tab:extensions}, we experiment with Gaussian noise with a strength of 0.2, PGD adversarial noise with $N=4$ \cite{Xie_Adversarial}, and AugMix \cite{hendrycks2019augmix}. While the use of a separated BatchNorm appears to yield improvements in all models, the best performing result still arises from the application of Weak Augment on top of RandAugment. This is perhaps not so surprising as RandAugment has the highest baseline score of all tested models, but Weak Augment appears to be a generic strategy that can yield improvements across different data augmentation types. 

\begin{table}
    \centering 
    \begin{tabular}{l|r|r}
        \multicolumn{3}{c}{\shortstack{\textbf{Clean Test Accuracy for Alternate Augmentation Strategies} \\ \textbf{Applied to the Auxliary BatchNorm}}} \\
        \hline
        \textbf{CIFAR-10} & Baseline & Weak Augment \\
        Gaussian ($\sigma$ = 0.2) & 93.2 & (+2.0\%) 95.2 \\
        Adversarial Noise & 94.7 & (+0.5\%) 95.2 \\
        AugMix & 94.8 & (+0.9\%) 95.7 \\
        RandAugment & 95.8 & (+0.5\%) 96.3 \\ 
        \multicolumn{2}{c}{} \\ 
    \end{tabular}
    \caption{Improvement from applying the Weak Augment training strategy for alternate data augmentation methods, compared to baseline models using the augmentation strategy and having only a single BatchNorm. For all of these studies, we provide the result for WideResNet-28-2 models trained on CIFAR-10 data. Weak Augment on top of RandAugment is shown to have the best performance, but the separated BatchNorm setup appears to have benefits for other trained models as well.}
    \label{tab:extensions}
\end{table}

\section{Separate BatchNorms Trade-off Between Accuracy and Robustness}
\label{sec:robustness}
Although impressive gains in test-set performance can be achieved from the Weak Augment setup from a separated BatchNorm, optimizing for accuracy naively introduces a new problem in these models: robustness. Table \ref{tab:robustness} presents the performance of the new Weak Augment models on the Common Corruptions benchmarks of CIFAR-10-C and ImageNet-C \cite{Hendrycks_Benchmarking}. Both CIFAR-10 models display significantly diminished robustness results.

\begin{table}[t]
    \centering
    \begin{tabular}{l|r|r|r}
        \multicolumn{4}{c}{\textbf{Corruption Error}} \\
        \hline
        & Standard & RA & Weak Augment (Main BN) \\
        \hline
        \textbf{CIFAR-10-C (uCE)} & & \\
        Wide-ResNet-28-2 &  27.7  & 15.9 &  19.1 \\
        Wide-ResNet-28-10 & 24.4  & 13.2 & 17.1 \\ 
        \hline
        \textbf{ImageNet-C (CE)} & &  \\ 
        ResNet50 & 77.5 & 70.8 & 69.2  \\
        \multicolumn{4}{c}{} \\ 
    \end{tabular}
    \caption{Corruption error based on the CIFAR-10-C and ImageNet-C tasks. For CIFAR-10-C,the unnormalized corruption error with the mean of 10 runs is reported. For ImageNet, the corruptions errors provided are after scaling by those of AlexNet. Lower scores are better.}
    \label{tab:robustness}
\end{table}

\subsection{Averaging Predictions from Both BatchNorms Leads to Improvements}
\begin{figure}
    \centering
    \includegraphics[trim=4 0 0 0,clip,width=\linewidth]{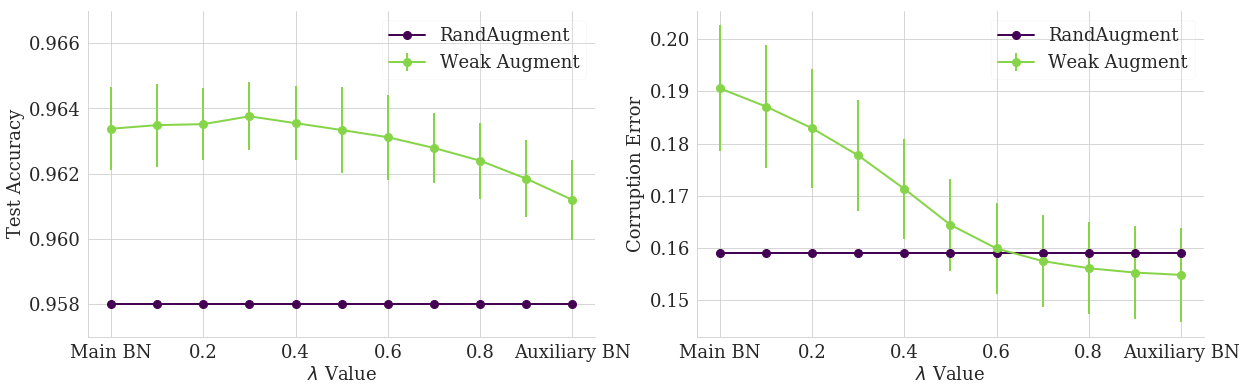}
    \caption{Effects of various $\lambda$ on the performance (accuracy and corruption error) of the Weak Augment setup. The figures show a clear trade-off between accuracy and robustness when using either the main or auxiliary BatchNorm ($\lambda = 0$ or $\lambda = 1$). However, at intermediate values such as $\lambda=0.5$, the model shows impressive gains in test accuracy without sacrificing model robustness. Note, higher accuracy is better whereas lower corruption error is better.}
    \label{fig:interpolate}
\end{figure}

While the main BatchNorm branch of the models appear to be more sensitive to input noise, this problem can be overcome by also incorporating the predictions coming from the auxiliary BatchNorm. Figure \ref{fig:interpolate} explores the performance of predictions coming from using the main and auxiliary BatchNorms, also including combinations thereof weighted by a factor of $\lambda \in [0,1]$. 

This graph presents a fundamental trade-off between accuracy and robustness between the predictions coming from the two BatchNorms, yet a simple average shows overall improvement in both metrics. The same holds true for the ImageNet model, where simple averaging yields a improvement of $+0.2\%$ in top-1 accuracy and -2.4 in CE robustness (lower is better). This suggests that utilizing both BatchNorms for evaluation may be a promising method for countering diminished robustness properties, if one is willing to compute both at inference. We hope that future work explores this direction futher, possibly determining if there is a way to distill the information in both BatchNorms into a single forward pass.

\subsection{Fourier Sensitivity Provides Another Perspective of Robustness}
\begin{figure}
    \centering
    \begin{subfigure}{0.3125\linewidth}
        \centering
        \includegraphics[width=0.6\textwidth]{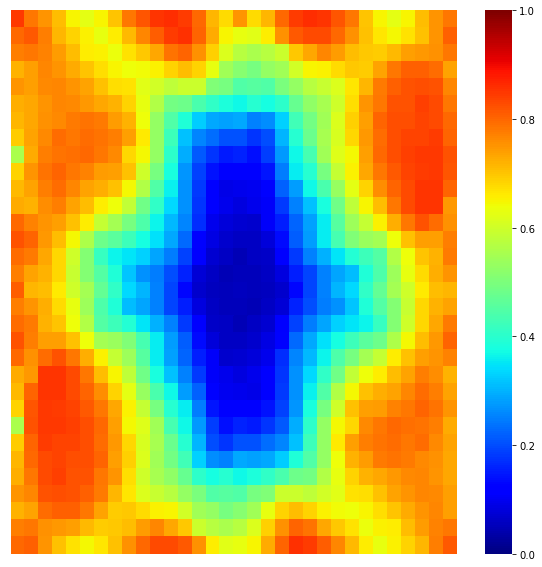}
        \caption{Main BatchNorm}
        \label{fig:heatmaps_no}
    \end{subfigure} %
    \begin{subfigure}{0.3125\linewidth}
        \centering
        \includegraphics[width=0.6\textwidth]{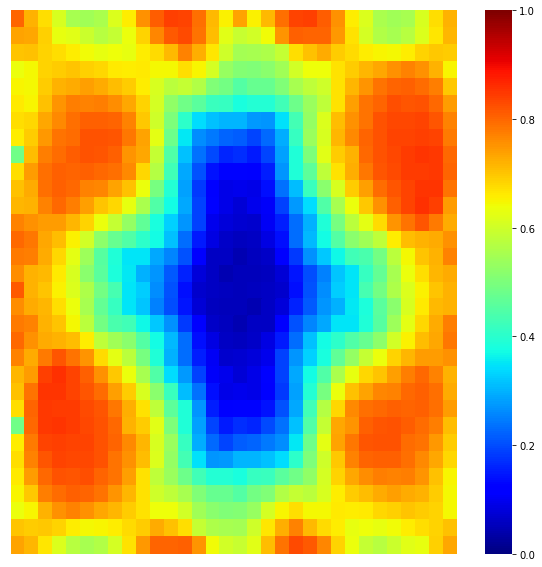}
        \caption{Average of Predictions}
        \label{fig:heatmaps_weak}
    \end{subfigure} %
    \begin{subfigure}{0.3125\linewidth}
        \centering
        \includegraphics[width=0.6\textwidth]{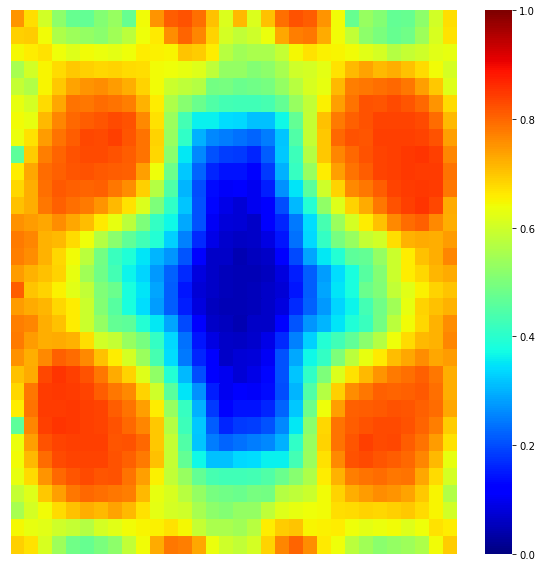}
        \caption{Auxiliary BatchNorm}
        \label{fig:heatmaps_rand}
    \end{subfigure}
    \caption{Heatmaps of the test error Fourier sensitivity of various models, using the method from \cite{Yin_Fourier}. Each figure shows the sensitivity to various sinusoidal gratings. We show the results for WideResNet-28-2 models trained on CIFAR-10, where we include the results for the main BatchNorm, the auxiliary BatchNorm, and the average of these two predictions ($\lambda = 0.5$). The corruption errors scale from 1.0 (red) to 0.0 (dark blue).
    }
    \label{fig:heatmaps}
\end{figure}

The robustness studies so far have only focused on a few parameterized noises from the Common Corruptions benchmark. In order to get a more general perspective on model robustness, we turn to analyzing noise in Fourier space using the strategy introduced by Yin et al. \cite{Yin_Fourier}. This evaluation strategy involves perturbing each image in the test set with noise sampled from different orientations and frequencies in Fourier space and then determining the difference in test set performance. These results are plotted in a heatmap, where the position indicates the noise in Fourier-space, with the lowest frequencies in the center of the images. The color then represents the test error. The results are presented in Figure \ref{fig:heatmaps}, specifically for the WideResNet-28-2 models trained on CIFAR-10. For these examples, the magnitude of the noise is set to have a $l_2$ norm to be 8.0, larger than in previous studies \cite{lopes2019improving, Yin_Fourier} but allowing for better visualization of the differences between models. Following the interpolation results presented in the previous section, the figures show the heatmaps for the main BatchNorm, the auxiliary BatchNorm, and the average of these two predictions.

The Frequency analysis showcases the same patterns as the Common Corruptions robustness in the previous section. Specifically, the main BatchNorm appears highly sensitive, particularly to high frequency noise. In contrast, the auxiliary BatchNorm ($\lambda = 1.0$) show greater robustness overall. Interpolating between the two predictions showcases similar benefits as before, maintaining most of the additional benefit gained from strong augmentation.

\subsection{Where are the improvements in robustness from separated BatchNorms coming from?}
\begin{figure}[ht!]
    \centering
    \includegraphics[width=0.8\textwidth]{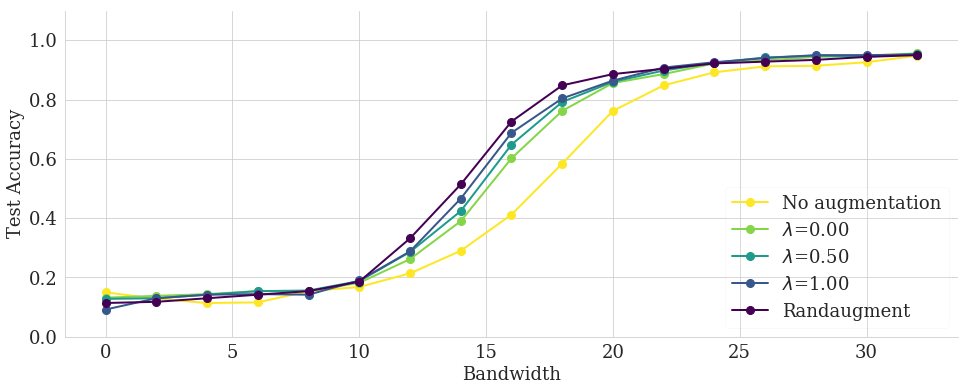}
    \caption{Effects of increasing the size of a low pass filter on the test performance of various WideResNet-28-2 models trained on CIFAR-10. The distributions show that RandAugment uses lower frequency information than a model with no augmentations. Our model, Weak Augment, appears to interpolate between these two regimes.}
    \label{fig:lowpass}
\end{figure}

Finally, we analyze which frequencies are most important for model performance and gain insights into why the Weak Augment method is effective. This is achieved by defining a low pass filter of bandwidth $B$ as the operation that all of the frequency components outside of a centered square of width $B$ in the Fourier spectrum, centered around the lowest frequency, to be zero. Then, an inverse Discrete Fourier Transform is applied to recover the image. Figure \ref{fig:lowpass} presents the effect of applying a low pass filter with increasing bandwidth on 500 examples from the CIFAR-10 test set and measures the difference in accuracy for WideResNet-28-2 models tested with no augmentations, Weak Augment of varying $\lambda$ parameters, and RandAugment. Generally, RandAugment uses significantly lower frequency information to achieve peak classifier performance when compared to no augmentations. The Weak Augment model effectively interpolates between these two regimes. The improvements in clean accuracy coming from Weak Augment could potentially be explained by the more effective use of higher frequency information available in the training images.

\section{Conclusion}
In this work, we analyze the popular training paradigm of using separated BatchNorm parameters and show that it can be applied to generic data augmentation setups. Specifically, we find that naively applying this approach and using clean (un-augmented) and augmented images to define the two BatchNorms does not lead to any improvements. Instead, we propose using \textbf{weak} augmentations rather than clean ones to define the main BatchNorm and find significant improvements on CIFAR-10, CIFAR-100, and ImageNet benchmarks. Finally, we find that defining BatchNorm parameters based on \textit{weak} augmentations leads to problems in model robustness but that this problem can be overcome by interpolating between the predictions of the two BatchNorm parameters.

\section*{Acknowledgements}
We thank Irwan Bello and the Google Brain Team for helpful feedback on this manuscript.

\newpage

\bibliographystyle{abbrv}
\bibliography{main}

\newpage

\appendix

\end{document}